%% file: main.tex
\documentclass[runningheads]{llncs}

\usepackage{eccv}

\usepackage{eccvabbrv}

\usepackage{graphicx}
\usepackage{booktabs}

\usepackage[accsupp]{axessibility}  %

\usepackage{multirow}
\usepackage{multicol}
\usepackage{wasysym}
\usepackage[dvipsnames]{xcolor}
\usepackage{makecell}
\usepackage{enumerate}

\input{premable}

\usepackage{hyperref}

\usepackage{orcidlink}

\begin{document}

\title{Robust Bird's Eye View Segmentation\\by Adapting DINOv2} 

\titlerunning{Robust BEV via DINOv2}

\author{Merve Rabia Barın\inst{1, 2}\orcidlink{0009-0001-3145-2670} \and
Görkay Aydemir\inst{1}\orcidlink{0009-0002-0315-3312} \and
Fatma G\"uney\inst{1, 2}\orcidlink{0000-0002-0358-983X}}

\authorrunning{Barın et al.}

\institute{Department of Computer Engineering, Koç University \and
KUIS AI Center}

\maketitle

\input{sections/0_abstract}    
\input{sections/1_intro}

\input{sections/2_rw}

\input{sections/3_methodology}

\input{sections/4_exp}

\input{sections/5_conc}

\section*{Acknowledgements}
This project is co-funded by KUIS AI and the European Union (ERC, ENSURE, 101116486). Views and opinions expressed are however those of the author(s) only and do not necessarily reflect those of the European Union or the European Research Council. Neither the European Union nor the granting authority can be held responsible for them.

\newpage
{
    \small
    \bibliographystyle{splncs04}
    \bibliography{main}
}

\end{document}

%% file: premable.tex
\newcommand{\bK}{\mathbf{K}}

\newcommand{\by}{\mathbf{y}}

\newcommand{\bV}{\mathbf{V}}
\newcommand{\bW}{\mathbf{W}}

\newcommand{\bq}{\mathbf{q}}

\newcommand{\bp}{\mathbf{p}}

\newcommand{\bB}{\mathbf{B}}

\newcommand{\bA}{\mathbf{A}}

\newcommand{\bff}{\mathbf{f}}

\newcommand{\bE}{\mathbf{E}}

\newcommand{\nR}{\mathbb{R}}

\newcommand{\boldparagraph}[1]{\vspace{0.2cm}\noindent{\bf #1:} }

\newcommand{\figref}[1]{Figure~\ref{#1}}
\newcommand{\secref}[1]{Section~\ref{#1}}

\newcommand{\tabref}[1]{Table~\ref{#1}}

\usepackage[dvipsnames]{xcolor}

%% file: sections/0_abstract.tex
\begin{abstract}
Extracting a Bird's Eye View (BEV) representation from multiple camera images offers a cost-effective, scalable alternative to LIDAR-based solutions in autonomous driving. However, the performance of the existing BEV methods drops significantly under various corruptions such as brightness and weather changes or camera failures. To improve the robustness of BEV perception, we propose to adapt a large vision foundational model, DINOv2, to BEV estimation using Low Rank Adaptation (LoRA). 
Our approach builds on the strong representation space of DINOv2 by adapting it to the BEV task in a state-of-the-art framework, SimpleBEV. Our experiments show increased robustness of BEV perception under various corruptions, with increasing gains from scaling up the model and the input resolution. We also showcase the effectiveness of the adapted representations in terms of fewer learnable parameters and faster convergence during training.
\end{abstract}

%% file: sections/1_intro.tex
\section{Introduction}
\label{sec:intro}

Accurate perception of the surrounding scene in 3D is crucial for safe navigation in autonomous driving. LIDAR sensors can provide accurate 3D measurements, however, due to their high cost and consequently scalability issues, camera-based solutions are explored as an alternative.
Specifically, substantial efforts have been directed toward extracting bird's-eye view (BEV) representations from multi-camera images~\cite{Jonah2020ECCV, Brady2022CVPR, Harley2023ICRA, Zhiqi2022ECCV, Loick2024CVPR}, providing a more cost-effective solution. BEV representations are assumed as input to motion prediction methods~\cite{Chang2019CVPR, Holger2020CVPR} and are increasingly used as part of end-to-end driving systems~\cite{Chen2019CORL, Zhang2021ICCV, Hanselmann2022ECCV, Chen2022CVPR, Hu2023CVPR}. 

While the robustness of these models under various conditions is a critical factor for ensuring safety, recent work~\cite{Xie2023ARXIV} shows that BEV perception models suffer from performance degradation when exposed to different types of corruption such as brightness changes, adversarial weather conditions, motion blur, quantization, frame loss, and camera crash, highlighting a significant challenge. The accuracy of BEV perception plays a crucial role in both motion prediction~\cite{Xu2024ICRA} and end-to-end driving. While privileged agents that have access to ground truth BEV demonstrate an impressive driving performance, their student counterparts suffer from mistakes in predicted BEV maps~\cite{Zhang2021ICCV}. Similarly, the performance of motion prediction methods drops notably while switching from ground truth BEV to the predicted BEV~\cite{Xu2024ICRA}.

The availability of large-scale datasets has facilitated the emergence of visual foundation models, renowned for their generalization capabilities. Notably, DINOv2~\cite{Oquab2023ARXIV} stands out for its robust, general-purpose visual features, making it a suitable choice for various tasks such as zero-shot correspondence estimation~\cite{Zhang2023NeurIPS}, robotics~\cite{Blumenkamp2024CORL}, object-centric learning~\cite{Aydemir2023NeurIPS}, point tracking~\cite{Aydemir2024ARXIV}, and object segmentation~\cite{Nguyen2023Cnos}. Despite the rich representation capacity of DINOv2, only a few works~\cite{Jonas2024IROS, Sirko-Galouchenko2024CVPRW} have explored its potential for BEV segmentation. %
In this work, we aim to explore the effectiveness of DINOv2 for robust BEV perception including performance, parameter efficiency, and convergence behavior.

\input{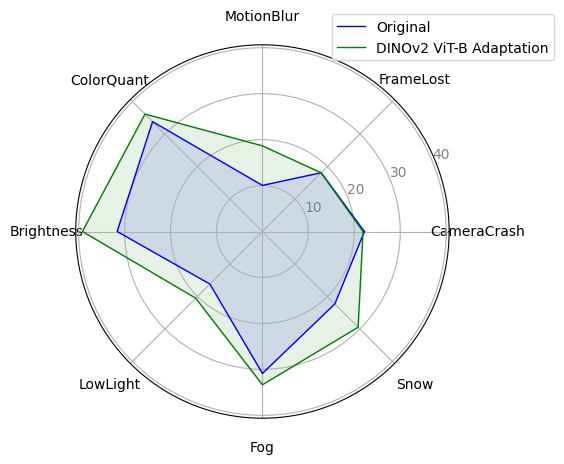}

We integrate DINOv2 into a state-of-the-art BEV segmentation model, SimpleBEV~\cite{Harley2023ICRA}, by utilizing an efficient adaptation technique~\cite{Hu2022ICLR}. Specifically, we replace the backbone of SimpleBEV with DINOv2 for feature extraction and then efficiently update it using Low Rank Adaptation (LoRA). We systematically analyze the effectiveness of our approach by comparing the robustness of our adaptation to the original SimpleBEV with ResNet-101 backbone in terms of accuracy, parameter efficiency, and convergence behavior. Our experiments reveal that our adaptation improves the robustness of BEV perception under adversarial conditions with significantly lower learnable parameters and shorter training times.

%% file: figures/robustness.tex
\begin{figure}[t]
    \centering
    \begin{subfigure}[b]{0.49\textwidth}
        \centering
        \includegraphics[width=1.0\textwidth]{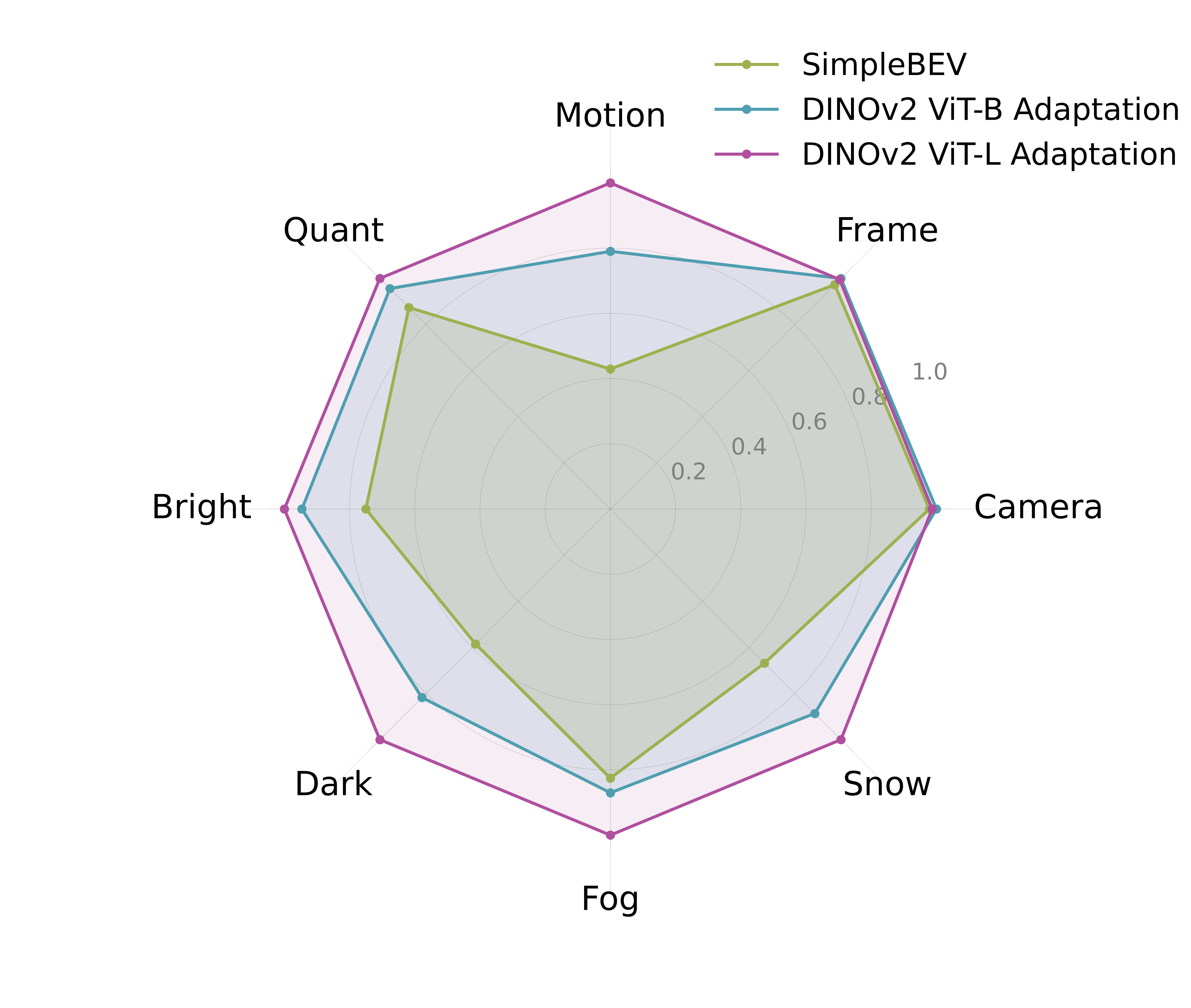}
        \caption{Performances under corruptions}
        \label{fig:robust_corr}
    \end{subfigure}
    \hfill
    \begin{subfigure}[b]{0.49\textwidth}
        \centering
        \includegraphics[width=1.0\textwidth]{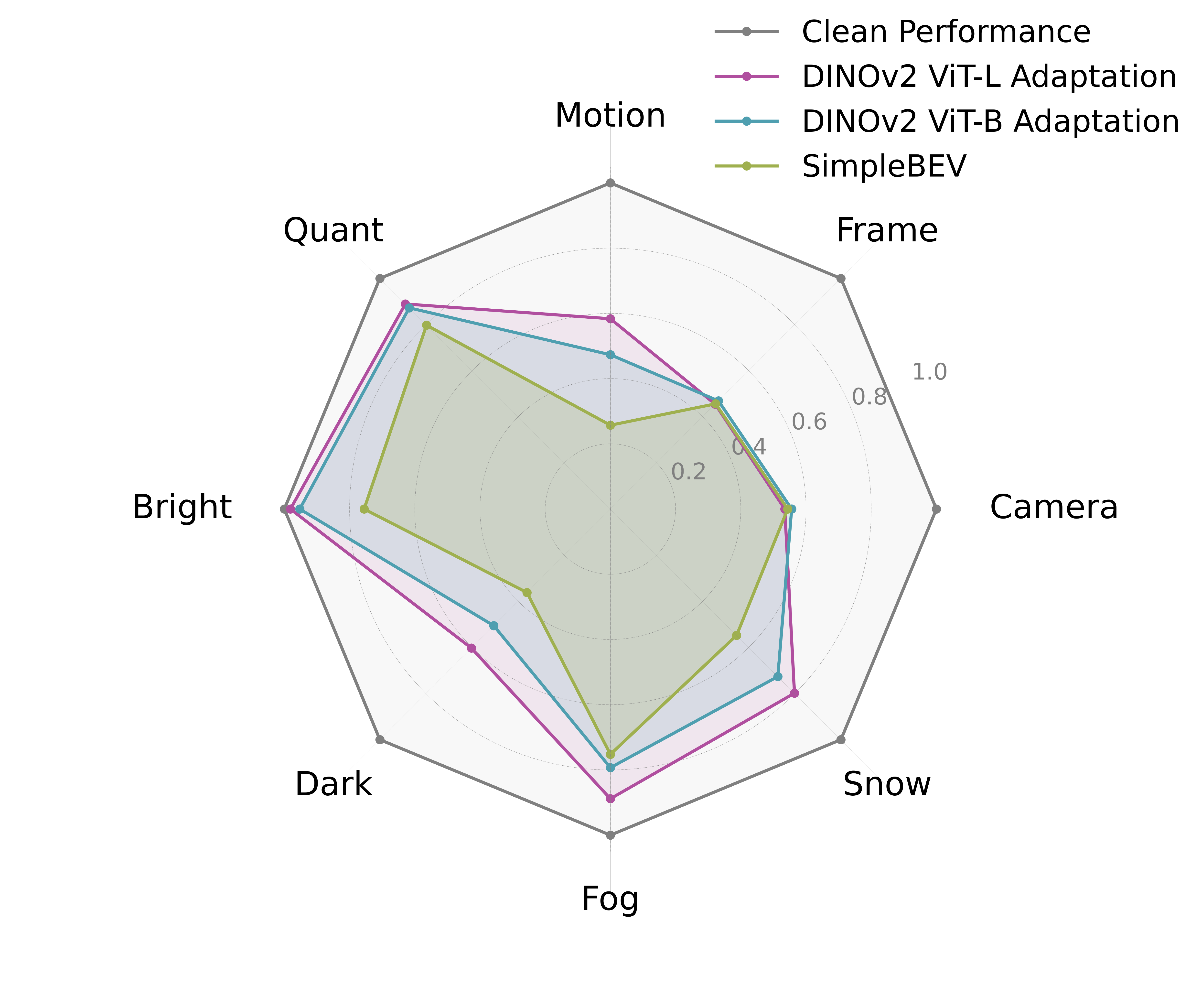}
        \caption{Performance Drop under Corruptions}
        \label{fig:robust_drop}
    \end{subfigure}
    \caption{\textbf{Robustness Analysis on nuScenes-C}. We compare the models under different types of corruptions in \subref{fig:robust_corr}, where each axis is normalized over the maximum performing model, \ie ViT-L adaptation. We show the performance drop of models relative to their performance on clean data in \subref{fig:robust_drop}, where each axis is normalized to the clean data performance of the corresponding model.}
    
    \label{fig:robustness}
\end{figure}

%% file: sections/2_rw.tex
\section{Related Work}
\label{sec:rw}

\boldparagraph{Camera-based BEV Segmentation}
Bird's-eye-view (BEV) representation is frequently used in autonomous driving to capture spatial information effectively. BEV summarizes the scene in a clear and compact representation by extracting the necessary 3D information from 2D images. %
BEV methods first process camera images using a backbone, and then construct a discrete 3D representation of the surrounding scene by transforming 2D image features to 3D voxel grids. After constructing the 3D voxel grid, the features are decoded into a 2D BEV representation of the scene, after flattening the height dimension. %

We can categorize the existing work on BEV into three based on how they extract 3D information from 2D images: 
\begin{enumerate}[i)]
    \item \textit{Depth-based} methods \cite{Jonah2020ECCV, Yinhao2023AAAI}, learn an explicit depth distribution at each pixel. They perform a weighted sum over the depth probabilities to back-project image features to 3D. However, learning an explicit depth distribution can be challenging due to the complexity of accurately modeling the depth for every pixel. Furthermore, voxel grids can only accumulate sparse features, which magnifies projection errors.
    \item \textit{Attention-based} models \cite{Brady2022CVPR, Bartoccioni2023PMLR} learn to align image features with voxel grids with an attention mechanism, leveraging camera-aware position embeddings to implicitly learn depth. However, learning projection via attention causes performance issues in attention-based approaches. %
    \item \textit{Sampling-based} methods \cite{Harley2023ICRA, Zhiqi2022ECCV} address these issues by sending rays from voxel grids to the images and bilinearly sampling the intersecting image features. Changing the direction of sampling increases the density of features in the voxel grid, relying on the quality of the features rather than the accuracy of projection. 
\end{enumerate}
Due to these reasons, we choose a sampling-based method, SimpleBEV~\cite{Harley2023ICRA} for our analysis.

\boldparagraph{Vision Foundational Models in BEV} Foundational models trained on large-scale data provide robust representations, improving performance in various downstream tasks~\cite{Zhang2023NeurIPS, Blumenkamp2024CORL, Aydemir2023NeurIPS, Nguyen2023Cnos} due to their inherent semantic understanding and strong generalization capabilities. Although foundational models are trained solely on 2D data, they are shown to capture some 3D information from images, as tested on multi-view correspondence and depth estimation tasks~\cite{Zhan2023ARXIV, El2024CVPR}. 
Among the foundational models evaluated, DINOv2~\cite{Oquab2023ARXIV} performs the best for the 3D tasks considered, alongside Stable Diffusion~\cite{Rombach2021CVPR2022}, indicating its potential for 3D scene understanding. There is recent work building on foundational models for pre-training a BEV network for occupancy prediction~\cite{Sirko-Galouchenko2024CVPRW} or sensor fusion~\cite{Jonas2024IROS} by benefiting from semantic capabilities of DINOv2~\cite{Oquab2023ARXIV}. We directly target improving BEV estimation from camera images by adapting DINOv2.

%% file: sections/3_methodology.tex
\section{Methodology}
Our goal is to integrate visual foundation model DINOv2~\cite{Oquab2023ARXIV} into BEV prediction model, SimpleBEV~\cite{Harley2023ICRA}. In this section, we first explain our adaptation strategy, Low Rank Adaptation~(\secref{sec:prel1}) and then SimpleBEV~(\secref{sec:prel2}) for completeness, and finally, present our approach to integrate DINOv2 into the SimpleBEV framework~(\secref{sec:dinoforbev}).

\subsection{Low Rank Adaptation (LoRA)}
\label{sec:prel1}
Low Rank Adaptation~\cite{Hu2022ICLR}, \ie LoRA, is widely used in Natural Language Processing to adapt pretrained large models to different tasks. LoRA is parameter efficient compared to fine-tuning, as only low-rank matrices are trained as residuals to frozen weights. Following the approach in MeLo~\cite{Zhu2023ARXIV}, we update only the query and value projections in all attention layers in the ViT. Formally, given a pre-trained weight matrix $\bW_{\{Q, V\}} \in \nR^{d \times d}$, we obtain $\bW'$ as the result of adaptation:
\begin{equation}
    \bW'_{\{Q, V\}}= \bW_{\{Q, V\}} + \bB \bA 
\end{equation}
where $\bA \in \nR^{r \times d}$ and $\bB \in \nR^{d \times r}$ are learned matrices, $r$ is the rank, and $d$ is the feature dimension. We train only the $\bB$ and $\bA$ matrices while keeping the original $\bW$ frozen for each attention layer.

\subsection{SimpleBEV}
\label{sec:prel2}
Given images from $N$ cameras, SimpleBEV~\cite{Harley2023ICRA} first extracts a feature map $\bff_i \in \nR^{d \times H_f \times W_f}$ for each image $i \in \{1, 2, \dots, N\}$, where $d$ represents the feature dimension, and $H_f \times W_f$, the size of the feature map. 
The method then employs a parameter-free lifting technique to transform image features into a 3D voxel grid $\bV \in \nR^{d \times X \times Y \times Z}$, where $X$, $Y$, and $Z$ correspond to the width, height and depth of the grid, respectively.

For each voxel $\bV_\bp \in \nR^d$ in the grid, represented by the 3D coordinate $\bp = [x, y, z]$, the corresponding 2D pixel coordinates $\bq_i = (u_i, v_i)$ is calculated by projecting $\bp$ onto the 2D image plane of each camera $i$, using the intrinsic and extrinsic matrices $\bK_i$ and $\bE_i$:
\begin{equation}
    \bq_i = \bK_i \bE_i \bp
\end{equation}
The feature values are then bilinearly sampled from the feature map $\bff_i$ at these projected coordinates, $\bq_i$. Then, sampled features from all $N$ cameras are aggregated and assigned to the corresponding voxel on the 3D grid:
\begin{equation}
    \bV_\bp = \frac{1}{N} \sum_{i=1}^{N} \text{sample}(\bff_i, \bq_i)
    \label{eq:bev_sample}
\end{equation}
After aggregating features for all $\bV_\bp \in \bV$ and constructing a 3D voxel grid $\bV$ that encapsulates the entire scene, a compressor reduces this voxel grid into a 2D BEV feature map $\bB \in \nR^{d \times X \times Z}$. The decoder then predicts the probability of occupancy for each grid cell. The model is trained using binary cross-entropy loss to optimize these predictions.

\input{figures/overview}

\subsection{Adaptation of DINOv2 to BEV}
\label{sec:dinoforbev}

\boldparagraph{Integrating DINOv2} We perform two main steps to integrate DINOv2 into SimpleBEV. First, we introduce additional layers in the query and key components of the ViT-based DINOv2 for adaptation, as described in \secref{sec:prel1}. Next, we replace the feature map $\bff_i \in \mathbb{R}^{d \times H_f \times W_f}$ with the output tokens of DINOv2 for each corresponding camera view, followed by pooling as shown in~\eqref{eq:bev_sample}. During training, only the additional layers are updated, while the original DINOv2 model remains frozen. SimpleBEV, with its original ResNet-101 backbone, serves as the baseline for comparison. 

\boldparagraph{Evaluation Aspects} We assess the quality of our adaptation by focusing on three key aspects in the evaluation:
\begin{enumerate}[i)]
    \item First, we consider \textbf{input resolution}, both image and feature map, where the ability to achieve high performance with lower-resolution inputs is advantageous, as it demonstrates efficiency in processing. 
    \item Second, we evaluate the \textbf{number of learnable parameters}, which not only indicates the resource efficiency of training but also reflects the robustness and general-purpose nature of the input features. Models with fewer parameters that still perform well suggest that the features are inherently strong, as they require minimal transformation.
    \item Lastly, we examine \textbf{convergence speed}, favoring models that reach optimal performance quickly, as this reduces computational and time-related costs during training. Convergence is measured by the number of updates needed, with faster convergence indicating more effective use of general-purpose features. 
\end{enumerate}
As a result, we prioritize models that excel in using lower-resolution inputs, have fewer learnable parameters, and converge more quickly, as these traits highlight the efficiency and robustness of our adaptation.

%% file: figures/overview.tex
\begin{figure}[h]
    \centering
    \includegraphics[width=1.0\textwidth]{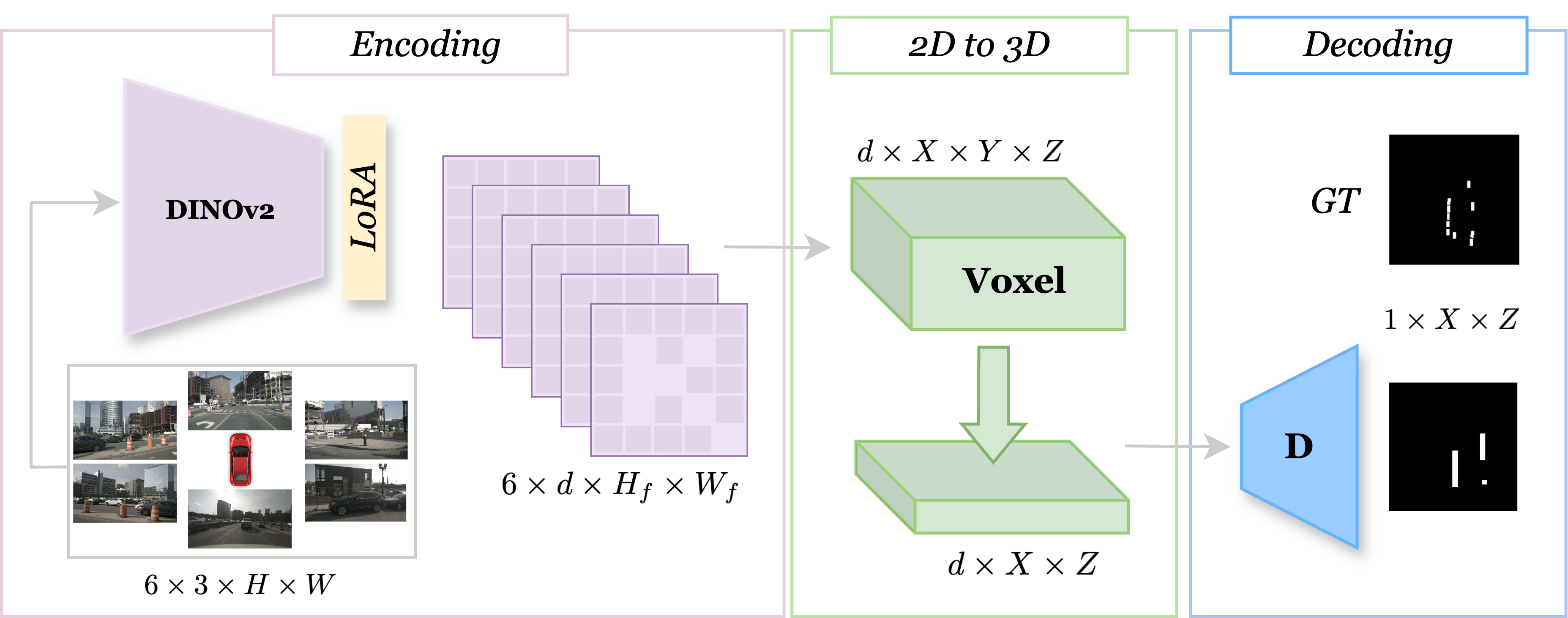}
    \caption{\textbf{Overview.} In this work, we propose to adapt DINOv2 to BEV segmentation using Low-Rank Adaptation (LoRA) for a robust BEV model. There are three main steps: i) We encode the camera images using DINOv2 to obtain tokens for each view, with attention weights updated through LoRA. ii) Transform image features from 2D to 3D using pull mechanism proposed by \cite{Harley2023ICRA}. iii) Decode BEV features to 2D vehicle BEV masks. }
    \label{fig:overview}
\end{figure}

%% file: sections/4_exp.tex
\input{tables/comparison}

\section{Experiments}
\subsection{Experimental Setup}

\boldparagraph{Datasets and Metrics} We conducted our experiments on the nuScenes~\cite{Holger2020CVPR} dataset, which is widely used for training and evaluating camera-based BEV segmentation methods. The dataset contains 28130 instances in the training set and 6019 instances in the validation set. %
For robustness analysis, we conducted experiments on the nuScenes-C benchmark~\cite{Xie2023ARXIV}, which is designed to measure the robustness of camera-based BEV perception models across eight types of corruptions under three levels of severity. The corruptions in this dataset are grouped into 8 categories: \textit{brightness, darkness, fog, snow, motion blur, color quantization, camera crash,} and \textit{frame loss}. For detailed descriptions of these augmentations, please refer to RoboBEV~\cite{Xie2023ARXIV}. 
Following prior work, we use the mean Intersection-over-Union (mIoU) to evaluate model performance, measuring the overlap between our predictions and the ground truth boxes.

\boldparagraph{Training Details} We train the adaptation models using ViT-B and ViT-L architectures with the AdamW optimizer, a learning rate of  $1 \times 10^{-3}$, and one cycle scheduler. For fine-tuning, we use a lower learning rate, $1 \times 10^{-5}$, and set the effective batch size to 16. %
For the ViT-B model, we followed the default training schedule of 25K updates as suggested in \cite{Harley2023ICRA}. For the ViT-L model, we additionally explored shorter training time, using approximately one-third of the default steps to demonstrate the effect of adaptation on convergence with a large backbone. 

\subsection{Adaptation} 
\label{subsection:adaptation}
We compare the performance of our adaptation models to the original SimpleBEV results in \tabref{tab:comparison}. Our comparisons focus on three key aspects: i) Same input resolution, ensuring the models are evaluated under identical conditions; ii) Same feature map resolution to assess the efficiency of the spatial features encoded by the models; iii) Update iterations to understand the convergence speed and training efficiency of the models.

\boldparagraph{Same Input Resolution} Vision-based BEV perception models rely entirely on image input, making image resolution critically important. As demonstrated in~\cite{Harley2023ICRA}, increasing the input resolution can improve performance up to a certain limit. Considering these, we choose to experiment on $224 \times 400$ and $448 \times 800$. We compare the models with nearly identical input resolutions\footnote{The input resolution for ViTs should be divisible by the patch size, which is 14 in the case of DINOv2. Therefore, we use the closest multiples of 14 to match the target resolution.} to assess their effectiveness when processing a similar number of pixels for a fair evaluation. %
For the smaller resolution, SimpleBEV and DINOv2 ViT-B reach the same level of performance, 42.3 mIoU (A \vs C), while ViT-L surpasses SimpleBEV by 1.1 points (A \vs D). For the higher resolution, ViT-L outperforms SimpleBEV by a small margin, with mIoUs of 47.4 and 47.7, respectively (B \vs G). This demonstrates that the adapted DINOv2 backbones can reach the performance of SimpleBEV using similar resolutions.

\boldparagraph{Same Feature Resolution} In addition to input resolution, we also consider feature map resolution. By this, we refer to the resolution of $\bff_i$ as introduced in \secref{sec:prel2}, which is the output of the backbone. Different backbones operate with different strides, \ie the downsampling ratio of the final feature map. A backbone with a lower stride is expected to be advantageous~\cite{Karaev2024ECCV}, as it can better preserve the details of spatial information. Specifically, the SimpleBEV downsamples the image to $  1/8^{th} $ of the original input resolution, while DINOv2 downsamples to $  1/{14}^{th} $ due to patch size. Considering the same feature resolution of $28 \times 50$, the adapted ViT-L outperforms SimpleBEV by a significant margin of 5.3 IoU (A \vs F). %
This indicates that the adapted DINOv2 backbone can preserve spatial information more efficiently.

Moreover, increasing the feature resolution consistently improves performance, as expected, by providing more accurate and dense interpolation among features. For instance, raising the resolution from $16 \times 28$ to $28 \times 50$ results in a 4.2 mIoU increase (D \vs F). Additionally, a further resolution increase to $32 \times 56$ yields a slight improvement, raising the mIoU from 47.6 to 47.7 (F \vs G).

\boldparagraph{Number of Updates} We chose the DINOv2 ViT-L model for convergence experiments due to its large scale, making it ideal for assessing convergence efficiency. The performance of DINOv2 ViT-L shows a drop of 0.2 IoU when trained only for one-third of the total iterations (D \vs E). In contrast, ViT-L not only surpasses SimpleBEV but does so even with fewer training iterations, given the same input resolution (A \vs E) and the same feature resolution (A \vs F). These results indicate that DINOv2 adaptation can converge quickly, and shorter training times have minimal impact on performance.

\input{figures/training_rank_combined}

\subsection{Robustness Evaluation}
In~\figref{fig:robustness}, we present a robustness analysis comparing the SimpleBEV baseline with our two adaptation variants, ViT-B and ViT-L. We evaluate the models under various corruptions from the nuScenes-C dataset and report their performance separately to assess how each model handles different types of corruptions. For a fair comparison, all models are trained at similar resolutions: $224 \times 400$ for SimpleBEV and $224 \times 392$ for the ViT backbones, as discussed in \secref{subsection:adaptation}.

In \figref{fig:robust_corr}, the performance of all methods is normalized based on the best-performing model for each type of corruption. For color quantization, frame loss, and camera crash, all methods perform similarly. However, the ViT-L adaptation significantly outperforms other methods in most corruption types, surpassing SimpleBEV by at least 20\%, with ViT-B also showing strong results. The difference is particularly pronounced in the case of motion blur, where SimpleBEV operates at only 40\% of the ViT-L adaptation's performance. %

\figref{fig:robust_drop} illustrate the performance drop of each model relative to its performance on clean data. This figure highlights the relative decline in performance for each model when exposed to various corruptions. As seen in the previous analysis, the ViT-B and ViT-L adaptations exhibit less performance degradation compared to SimpleBEV. Excluding the camera crash and frame loss scenarios, the ViT-L adaptation consistently maintains its performance, never dropping below 60\% of its clean performance, with degradation of less than 20\% for brightness, fog, and quantization. For the ViT-B adaptation, the threshold is around 45\%. In contrast, SimpleBEV struggles to maintain its performance, with drops below 40\% in motion blur conditions and below 30\% in darkness. Overall, the DINOv2 adaptations demonstrate greater robustness than SimpleBEV in six out of eight corruption types and are comparable in the remaining two, camera crash and frame loss, highlighting the superior robustness of the adaptation approach. This finding underscores the value of exploring foundational models like DINOv2 for enhancing robustness in BEV perception tasks.

\subsection{Ablation Study}

\boldparagraph{Training Method} To highlight the impact of LoRA, we conducted experiments with the ViT-B DINOv2 adaptation across three configurations, as shown in \tabref{tab:adaptation}: i) Frozen, where only the decoder (with 5M parameters) is trained and no additional learnable parameters are introduced, \ie no updates to the backbone; ii) Fine-tuning, where all 86M parameters of the ViT-B DINOv2 backbone are updated; and iii) LoRA, where a small set of parameters is learned within the attention layers of the backbone. 

The experiments reveal that, while the frozen model demonstrates a decent zero-shot representation performance, it significantly lags behind SimpleBEV (34.3 \vs 42.3). Fine-tuning the entire backbone offers improved results but demands significant computational resources due to the large number of learnable parameters (86M). In contrast, the LoRA configuration, which adds just 1M parameters to the decoder, achieves results that are on par with (ViT-B; 42.3) or even superior (ViT-L; 43.4) to SimpleBEV (42.3). This is achieved by updating only 1.12\% of the parameters in ViT-B and 2.70\% in ViT-L. Notably, LoRA outperforms the full fine-tuning by 0.8 points, showcasing its parameter efficiency and effectiveness in enhancing the performance.

\boldparagraph{Rank of LoRA}
We experimented by varying the rank of adaptation in LoRA as shown in \figref{fig:lora_rank}. Rank essentially controls the capacity of the adaptation, with higher ranks providing more parameters for fine-tuning. Higher ranks can capture more complex relationships and lead to better performance, while potentially losing the information from pre-training. A rank of 0 corresponds to no updates to the backbone, \ie frozen backbone. We found that increasing the rank leads to consistent improvements up to a certain point, specifically up to rank 32. However, increasing the rank from 32 to 64 results in a performance decrease of 0.1. A potential reason for this is the loss of the inductive bias of DINOv2, due to the larger updates on the attention weights. This indicates that rank 32 strikes an optimal balance between adapting the features to the task and preserving the valuable information acquired during pre-training.

%% file: tables/comparison.tex
\begin{table*}[t]
  \centering
  \setlength{\tabcolsep}{6pt}
  \caption{\textbf{Quantitative Results on nuScenes Validation Set.} This table compares the results of SimpleBEV and DINOv2 adaptations with different backbones. The $\bullet\bullet\bullet$ represents the number of iterations in the full training of SimpleBEV, which corresponds to 25K steps, while $\bullet$ represents one-third of it. See text for details.}
\begin{tabular}{c ccccc c}
    \toprule
      \multirow{2}{*}{\textbf{Model}} & \multirow{2}{*}{\textbf{Backbone}} & \textbf{Input}  & \textbf{Feature} & \multirow{2}{*}{\#\textbf{Steps}} & \multirow{2}{*}{\textbf{mIoU}} & \multirow{2}{*}{\textbf{Exp.}} \\
     & & \textbf{Resolution} & \textbf{Resolution} & &  &  \\
    \midrule
    \multirow{2}{*}{SimpleBEV} & \multirow{2}{*}{ResNet-101} & 224 $\times$ 400 & $28 \times 50$ &                                    \multirow{2}{*}{$\bullet\bullet\bullet$}   & 42.3 & A   \\  
    & & 448 $\times$ 800 & $56 \times 100$ &  & 47.4 & B \\ 
    \midrule                 
    \multirow{5}{*}{DINOv2} &  ViT-B & \multirow{2}{*}{224 $\times$ 392} & \multirow{2}{*}{$16 \times 28$} & \multirow{2}{*}{$\bullet\bullet\bullet$}   & 42.3 & C   \\
     &  ViT-L & & &  &  43.4 & D  \\
     \cmidrule{2-7} 
     & \multirow{3}{*}{ViT-L} & 224 $\times$ 392 & $16 \times 28$ &  \multirow{3}{*}{$\bullet$} & 43.2 & E \\
     & & 392 $\times$ 700 & $28 \times 50$ & &  47.6 & F  \\
     & & 448 $\times$ 784 & $32 \times 56$ & & 47.7 & G  \\
    \bottomrule
  \end{tabular}
  \label{tab:comparison}
\end{table*}

%% file: figures/training_rank_combined.tex
\begin{figure}[!htb]
    \centering
    \begin{minipage}{0.49\textwidth}
        \centering
        \captionof{table}{\textbf{Training Method and Parameter Efficiency.} This table shows the results of different weight update strategies with the corresponding number of learnable parameters. Note that there are additional 5M parameters for the decoder.}
        \begin{tabular}{lcc c}
            \toprule
            \multirow{2}{*}{\textbf{Backbone}} &
            \multirow{2}{*}{\textbf{Method}} & \multirow{2}{*}{\textbf{\#Params}} &  \multirow{2}{*}{\textbf{mIoU}} \\
            & & & \\
            \midrule
            ResNet-101 & -       & 37M & 42.3 \\ 
            \midrule
            ViT-B & Frozen   & 0  &  34.3 \\
            ViT-B & Fine-tune & 86M & 41.5 \\  
            ViT-B & LoRA     & 1M  & 42.3 \\ 
            ViT-L & LoRA     & 3M & 43.4 \\ 
            \bottomrule
          \end{tabular}
    \label{tab:adaptation}
    \end{minipage}
    \hfill
    \begin{minipage}{0.45\textwidth}
        \centering
        \includegraphics[width=1.0\linewidth]{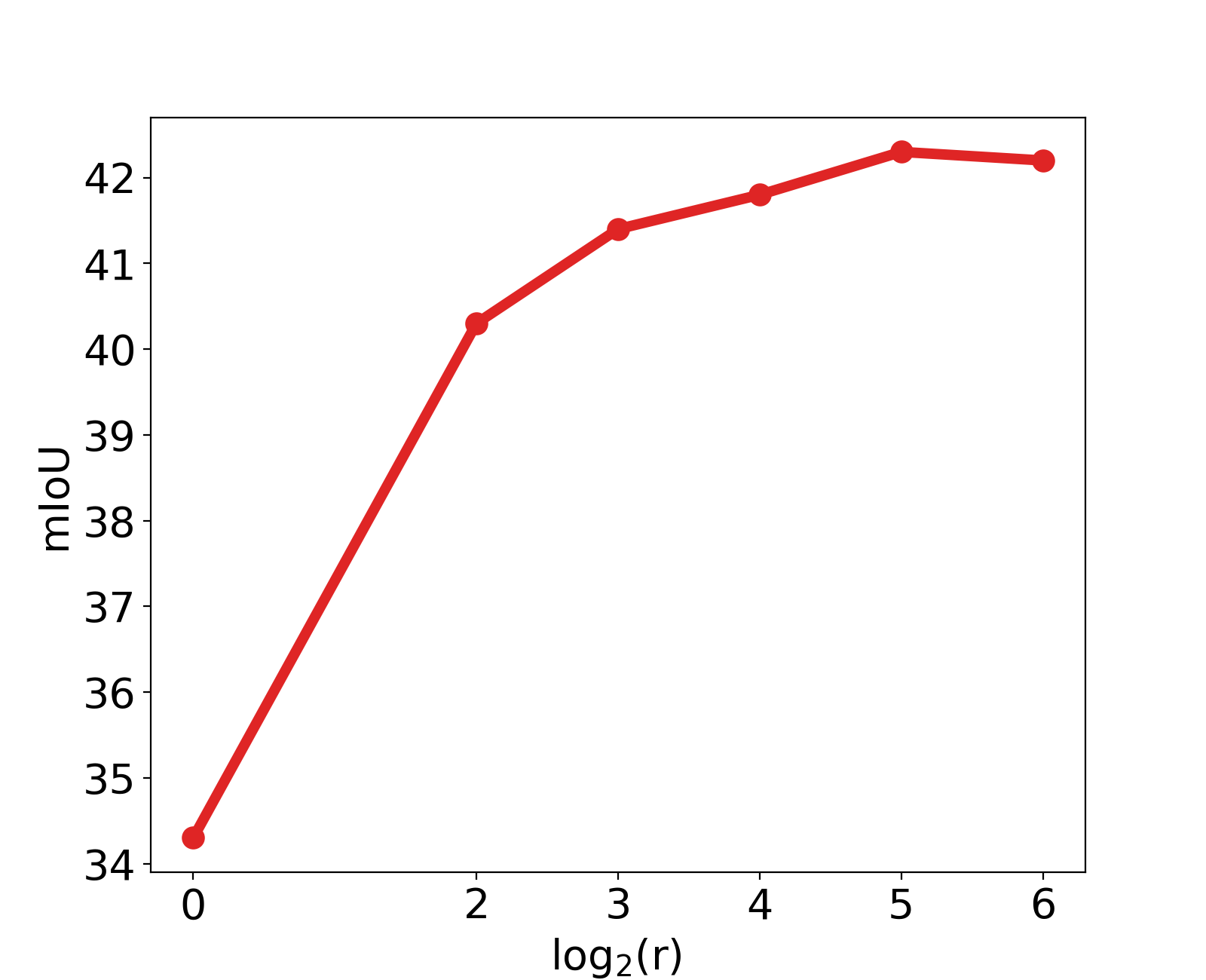}
        \caption{\textbf{Varying the Rank of LoRA.} This plot illustrates the effect of increasing the LoRA rank (in log scale) on the performance, with rank 0 representing a frozen backbone.}
        \label{fig:lora_rank}
    \end{minipage}
\end{figure}

%% file: sections/5_conc.tex
\section{Conclusion}
We investigated the effectiveness of DINOv2 with Low Rank Adaptation (LoRA) for BEV segmentation on nuScenes. We first showed comparable results to SimpleBEV with a smaller backbone in the clean setting. By scaling up the model or the input and feature resolutions, we could obtain significant improvements compared to the baseline performance. 
Our experiments on nuScenes with corruptions (nuScenes-C) show increased robustness to various corruptions, even with a shorter training time. These results justify our motivation to build on general-purpose features from large foundation models for improving BEV segmentation. 
Our approach requires updating significantly fewer parameters compared to full fine-tuning or the supervised baseline.

Our analysis is currently limited to DINOv2 and does not explore the performance of other promising 3D-aware foundation models, such as Stable Diffusion. Comparing different foundation models for BEV presents an interesting direction for further research, offering more insights about how to incorporate these models into BEV frameworks.